\documentclass[journal]{IEEEtran}
\IEEEoverridecommandlockouts

\usepackage{amsmath, amssymb, amsfonts}
\usepackage{algorithmic}
\usepackage{algorithm}
\usepackage{graphicx}
\usepackage{textcomp}
\usepackage{xcolor}
\usepackage{booktabs}
\usepackage{multirow}
\usepackage{cite}
\usepackage{bm}
\usepackage{hyperref}
\usepackage{subcaption}
\usepackage{siunitx}
\usepackage{url}
\usepackage{stfloats}  
\usepackage{tikz}
\usetikzlibrary{arrows.meta, positioning, calc, fit, backgrounds}

\newcommand{\R}{\mathbb{R}}
\newcommand{\norm}[1]{\lVert #1 \rVert}

\newcommand{\argmax}{\operatorname*{arg\,max}}

\begin{document}

\title{PROBE: Probabilistic Occupancy BEV Encoding
with Analytical Translation Robustness for \\ 3D Place Recognition}

\author{Jinseop~Lee, Byoungho~Lee, and Gichul~Yoo$^{*}$%
  \thanks{Manuscript received: March 5, 2026; Revised: May 4, 2026;
    Accepted: May 25, 2026.
    This paper was recommended for publication by
    Editor Lucia Pallottino upon evaluation of the Associate Editor
    and Reviewers' comments.
    $^{*}$Corresponding author.
    J.~Lee, B.~Lee, and G.~Yoo are with SK Intellix, Seoul, Republic of Korea
    (e-mail: \texttt{jinseop.llee@gmail.com, byoungho.lee1@sk.com,
    gichul.yoo@sk.com}).
    Digital Object Identifier (DOI): see top of this page.}%
}

\markboth{IEEE ROBOTICS AND AUTOMATION LETTERS. PREPRINT VERSION.
  ACCEPTED JUNE, 2026}%
{Lee \MakeLowercase{\textit{et al.}}: PROBE: Probabilistic Occupancy
  BEV Encoding with Analytical Translation Robustness
  for 3D Place Recognition}

\maketitle

\begin{abstract}
We present \textbf{PROBE} (\textbf{PR}obabilistic \textbf{O}ccupancy
\textbf{B}EV \textbf{E}ncoding), a learning-free LiDAR place recognition
descriptor that models each BEV cell's occupancy as a Bernoulli
random variable.
Rather than relying on discrete point-cloud perturbations, PROBE
analytically marginalizes over continuous Cartesian translations
via the polar Jacobian, yielding a distance-adaptive angular
uncertainty $\sigma_\theta = \sigma_t / r$ in $\mathcal{O}(R{\cdot}S)$ time.
The primary parameter~$\sigma_t$ represents the expected
translational uncertainty in meters, a sensor-independent
physical quantity that enhances cross-sensor generalization
{while reducing the need for extensive per-dataset tuning.}
Pairwise similarity combines a \emph{Bernoulli-KL Jaccard} with
exponential uncertainty gating and FFT-based height cosine
similarity for rotation alignment.
Evaluated on four datasets spanning four diverse LiDAR types,
PROBE achieves the highest accuracy among handcrafted
descriptors in multi-session evaluation
and competitive single-session performance
relative to both handcrafted and supervised baselines.
The source code and supplementary materials are available at
\url{https://sites.google.com/view/probe-pr}.
\end{abstract}


\section{Introduction}
\label{sec:intro}

\IEEEPARstart{P}{lace} recognition, the ability to re-identify
previously visited locations from sensor observations, is a critical
component of simultaneous localization and mapping
(SLAM)~\cite{thrun2005probabilistic}, enabling loop closure
detection, kidnapped-robot recovery, and multi-session map merging.
Among the various sensor modalities, 3D LiDAR is widely
adopted due to its illumination invariance and accurate
range measurements.

Existing LiDAR place recognition methods can be broadly categorized
into three families.  \emph{Handcrafted global descriptors}
project each scan into a compact representation, typically a
Bird's-Eye-View (BEV) polar
grid~\cite{kim2018scan,kim2022sc++,wang2020lidar,lu2023deepring}
or range-elevation bins~\cite{kim2024narrowing}, and resolve the heading
ambiguity via exhaustive circular shift or frequency-domain
matching.  These methods are lightweight and require no training.
\emph{Learning-based methods}~\cite{uy2018pointnetvlad,
chen2021overlapnet,ma2022overlaptransformer,komorowski2021minkloc3d}
extract global features via deep networks, achieving strong
generalization at the cost of GPU inference and training data.
\emph{Local feature methods}~\cite{yuan2023std,yuan2024btc}
detect keypoints and build geometric constraints for verification,
but require {explicit geometric verification and are less suited for}
standalone pairwise evaluation.

Despite their practical success, the handcrafted BEV family, to which PROBE
belongs, suffers from a fundamental limitation that
remains an open challenge.

\begin{figure}[!t]
  \centering
  \includegraphics[width=0.95\columnwidth]{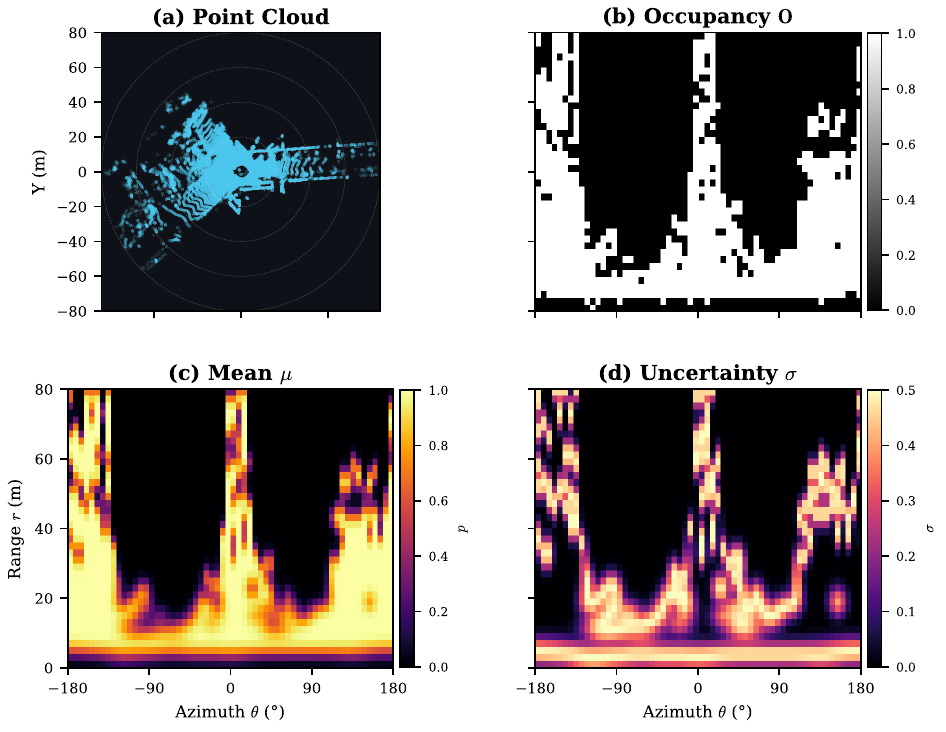}
  \caption{PROBE encodes a LiDAR point cloud (top-left)
    into a probabilistic polar BEV grid.  The binary
    occupancy mask~$\bm{O}$ (top-right), Bernoulli mean~$\mu$
    (bottom-left), and per-cell uncertainty~$\sigma$
    (bottom-right) are computed via Jacobian-derived analytical
    marginalization. High-$\sigma$ boundary cells are
    automatically downweighted during matching.}
  \label{fig:hero}
  \vspace{-2.7mm}
\end{figure}

\noindent\textbf{Binary occupancy and heuristic translation invariance.}
BEV polar grids are inherently sensitive to translational shifts:
a small lateral displacement of the sensor origin can toggle
boundary cells between occupied and empty, corrupting the
matching score.
SC++~\cite{kim2022sc++} mitigates this by constructing
descriptors at a small number of laterally shifted grid origins,
but covering only a discrete set of offsets leaves residual
sensitivity to arbitrary displacements.
Binary occupancy matching further exacerbates this issue by
treating all mismatched cells equally, without distinguishing
geometrically stable cells from volatile boundary cells that
toggle under slight viewpoint changes.

We propose \textbf{PROBE} (\textbf{PR}obabilistic \textbf{O}ccupancy
\textbf{B}EV \textbf{E}ncoding), which replaces
heuristic binary matching with a probabilistic model.
Instead of discrete spatial sampling, PROBE projects
isotropic Cartesian translation uncertainty into the polar domain
via the Jacobian.
This \emph{analytical marginalization} models each cell as a
Bernoulli random variable $(\mu, \sigma)$ within a single BEV grid
(Fig.~\ref{fig:hero}),
where the distance-adaptive uncertainty
$\sigma_\theta = \sigma_t / r$ follows directly from the
polar coordinate transform.
PROBE then employs a \emph{Bernoulli-KL Jaccard} that
downweights uncertain cells through exponential gating,
combined with FFT-accelerated cosine similarity for rotation
alignment.

The main contributions are:
\begin{enumerate}
  \item \textbf{Analytical Marginalization via Polar Jacobian}: We replace computationally expensive discrete point-cloud perturbations with a closed-form probabilistic model. By applying a Jacobian-derived adaptive 1D spatial blur, we analytically marginalize continuous Cartesian translations into a single BEV grid in $\mathcal{O}(R {\cdot} S)$ time, yielding a distance-adaptive angular uncertainty ($\sigma_\theta = \sigma_t / r$) and eliminating the need for generating multiple virtual views.
  \item \textbf{Bernoulli-KL Jaccard with Uncertainty Gating}:
    a pairwise scoring mechanism that smooths each
    cell's occupancy toward an uninformative prior proportionally
    to its uncertainty, computes symmetric KL divergence, and
    downweights high-$\sigma$ cells via exponential gating.
    This replaces the standard binary Jaccard with a divergence
    measure that distinguishes stable structures from
    viewpoint-sensitive boundaries.
  \item \textbf{Cross-sensor generalization with a single
    physically-grounded parameter}:
    the primary parameter~$\sigma_t$ represents
    the expected translational uncertainty in meters, a
    sensor-independent physical quantity that generalizes
    to unseen sensors and environments {while reducing the need
    for extensive per-dataset tuning}, as validated across 22 sequence configurations
    spanning four LiDAR types
    (KITTI, HeLiPR, NCLT, ComplexUrban).
\end{enumerate}

\begin{figure*}[!t]
  \centering
\resizebox{\textwidth}{!}{%
\begin{tikzpicture}[
  body/.style  = {fill={rgb,255:red,227;green,239;blue,245}},
  scorbg/.style= {fill={rgb,255:red,245;green,240;blue,232}},
  scorbdr/.style={draw={rgb,255:red,196;green,163;blue,90}},
  finbg/.style = {fill={rgb,255:red,214;green,234;blue,242}},
  gc/.style    = {color={rgb,255:red,61;green,61;blue,61}},
  bc/.style    = {color={rgb,255:red,46;green,110;blue,142}},
  bd/.style    = {draw={rgb,255:red,46;green,110;blue,142}},
  dbox/.style = {rectangle, rounded corners=2pt,
    bd, thick, body, minimum height=0.9cm, minimum width=1.9cm,
    align=center, inner sep=3pt, font=\small},
  obox/.style = {rectangle, rounded corners=4pt,
    bd, thick, dashed, body, fill opacity=0.4, text opacity=1,
    minimum height=1.0cm, minimum width=2.4cm,
    align=center, inner sep=4pt, font=\small},
  sbox/.style = {rectangle, rounded corners=3pt,
    scorbdr, thick, scorbg, minimum height=0.9cm, minimum width=2.4cm,
    align=center, inner sep=4pt, font=\small},
  fbox/.style = {rectangle, rounded corners=3pt,
    bd, very thick, finbg, minimum height=0.9cm, minimum width=2.3cm,
    align=center, inner sep=4pt, font=\small\bfseries},
  cop/.style = {circle, bd, thin, fill=white,
    minimum size=0.55cm, inner sep=0pt, font=\small},
  arr/.style  = {-{Stealth[length=4pt,width=3.5pt]}, thick, gc},
  darr/.style = {-{Stealth[length=4pt,width=3.5pt]}, thick, dashed, bc},
  lb/.style   = {font=\footnotesize, gc, align=center},
  sl/.style   = {font=\scriptsize, gc, align=center},
  stit/.style = {font=\small\bfseries\sffamily, gc},
]
\node[lb] (raw) at (2,0) {Point\\Cloud $\mathcal{P}$};
\node[obox, minimum width=3.2cm] (gop) at (5.5,0)
  {BEV Polar Grid\\{\scriptsize Max-Height $\bm{G}$, Occupancy $\bm{O}$}\\{\scriptsize(Sec.~III-A)}};
\node[dbox, minimum width=2.4cm] (rk) at (9.5,-1.7)
  {Retrieval Key\\$\bm{k}\!\in\!\mathbb{R}^{2R}$\\
   {\scriptsize $\bar{\bm{G}}\|\bar{\bm{\mu}}$
   \;(Sec.~III-C)}};
\node[obox, minimum width=4.0cm] (bern) at (5.5,-3.4)
  {\textbf{Bernoulli Occupancy}\\
   {\scriptsize Polar Jacobian blur $\to$ $(\mu, \sigma)$}\\
   {\scriptsize(Sec.~III-B)}};
\draw[arr] (raw.east) -- (gop.west);
\draw[arr] (gop.south) -- (bern.north);
\draw[arr] (gop.east) -- (9.5,0) -- (rk.north);
\draw[arr] (bern.east) -- (9.5,-3.4) -- (rk.south);
\begin{scope}[on background layer]
  \node[fill=blue!4, rounded corners=6pt,
        fit=(raw)(gop)(bern)(rk),
        inner sep=0.3cm, inner ysep=0.5cm] (gI) {};
  \node[stit, anchor=south] at (gI.north)
    {I.~Descriptor Generation};
\end{scope}
\def\Rx{13}
\node[dbox, minimum width=1.5cm] (kd) at (\Rx,0) {KD-tree\\Top-$K$};
\node[obox, minimum width=2.8cm] (fft) at ({\Rx+3.5},0)
  {\textbf{FFT Alignment}\\
   $\delta^*\!=\!\arg\max\mathrm{CC}[\delta]$\\
   {\scriptsize on $\bm{G}_m$, $\bm{G}_q$
   \;(Sec.~III-D)}};
\node[lb] (qP) at (\Rx,-1.6) {Query\\$\{\mu_q,\sigma_q\}$};
\node[dbox, minimum width=2.2cm] (aQ) at ({\Rx+3.5},-1.6)
  {Aligned Query\\{\scriptsize Circ.~Shift by $\delta^*$}};
\node[lb] (mP) at (\Rx,-3.4) {Map\\$\{\mu_m,\sigma_m\}$};
\node[sbox, minimum width=2.6cm] (kl) at ({\Rx+3.5},-3.4)
  {\textbf{Bernoulli-KL Jaccard}\\
   $\mathcal{J}_{KL}$\\
   {\scriptsize shrinkage-regularized}\\
   {\scriptsize(Sec.~III-D)}};
\node[cop] (mul) at ({\Rx+7},-3.4) {$\times$};
\node[fbox] (fin) at ({\Rx+9.5},-3.4)
  {$S_{\text{PROBE}}\!=\!\mathcal{J}_{KL}\!\cdot\!\mathcal{C}$\\[1pt]
   $d = 1 - S_{\text{PROBE}}$};
\draw[arr] (rk.east) -- (11.5,-1.7) -- (11.5,0) -- (kd.west);
\draw[arr] (kd.east) -- (fft.west);
\draw[arr] (fft.south) -- (aQ.north);
\draw[arr] (qP.east) -- (aQ.west);
\draw[arr] (aQ.south) -- (kl.north);
\draw[arr] (mP.east) -- (kl.west);
\draw[arr] (kl.east) -- (mul.west) node[midway,above]{\scriptsize $\mathcal{J}_{KL}$};
\draw[arr] (mul.east) -- (fin.west);
\draw[darr] (fft.east) -- ({\Rx+7},0) -- (mul.north) node[pos=0.5,right]{\scriptsize $\mathcal{C}$};
\begin{scope}[on background layer]
  \node[fill=orange!4, rounded corners=6pt,
        fit=(kd)(fft)(qP)(aQ)(mP)
            (kl)(mul)(fin),
        inner sep=0.3cm, inner ysep=0.5cm] (gII) {};
  \node[stit, anchor=south, color=orange!80!black] at (gII.north)
    {II.~Pairwise Matching \& Scoring};
\end{scope}
\end{tikzpicture}%
}
  \caption{PROBE pipeline.
    \textbf{(I)}~Descriptor generation: a BEV polar
    grid~$\bm{G}$ is built with max-height encoding and occupancy mask~$\bm{O}$.
    Analytical marginalization
    via the polar Jacobian produces per-cell Bernoulli occupancy~$(\mu,\sigma)$.
    A rotation-invariant ring-mean key~$\bm{k}\in\mathbb{R}^{2R}$ is formed for
    KD-tree pre-filtering.
    \textbf{(II)}~Pairwise scoring: FFT-based rotation alignment on the
    max-height grids yields~$\delta^*$; the query is circularly shifted to
    produce an aligned query, which is scored against the map via a
    shrinkage-regularized Bernoulli-KL Jaccard
    $\mathcal{J}_{KL}$.  The final score fuses
    $\mathcal{J}_{KL}$ with the height cosine similarity
    $\mathcal{C}=\mathrm{CC}[\delta^*]$ multiplicatively:
    $S_{\text{PROBE}} = \mathcal{J}_{KL} \cdot \mathcal{C}$.}
  \label{fig:pipeline}
  \vspace{-2.7mm}
\end{figure*}

\section{Related Work}
\label{sec:related}

\subsection{Handcrafted Global Descriptors}
The BEV polar grid is a standard representation for
lightweight LiDAR place recognition.
M2DP~\cite{he2016m2dp} projects the cloud onto multiple 2-D planes at
varying azimuths and elevations and applies SVD to produce a compact
global descriptor that aims for rotation invariance,
though its fixed projection geometry limits adaptability to varying sensor configurations.
Scan Context~(SC)~\cite{kim2018scan} projects each scan into an
$R\!\times\!S$ grid of maximum heights and aligns two descriptors by
exhaustively shifting columns, selecting the azimuth with the lowest
cosine distance.
LiDAR-Iris~\cite{wang2020lidar} applies a Fourier transform to achieve
rotation invariance, and binarizes features extracted via LoG-Gabor
filtering for fast Hamming-distance matching.
SC++~\cite{kim2022sc++} accelerates retrieval via
ring and sector keys and introduces a Cartesian Context that
discretizes the BEV on a Cartesian grid, enabling linear-shift
alignment that is more robust to translational offset.
RING++~\cite{lu2023deepring} applies Radon and Fourier transforms to
multi-channel BEV features, producing roto-translation invariant
descriptors.
SOLiD~\cite{kim2024narrowing} reweights range-elevation bins to
handle restricted field-of-view scenarios on heterogeneous platforms.

Among the above descriptors, most do not explicitly address
translational offset.
SC++~\cite{kim2022sc++} is a notable exception, synthesizing
laterally shifted copies of the descriptor to cover plausible
displacements, though at increased matching cost.
RING++~\cite{lu2023deepring} achieves roto-translation
invariance through spectral transforms, yet its strong
invariance can match geometrically similar but spatially
distant places, reducing recognition precision.
PROBE addresses these limitations by replacing heuristics
with analytical marginalization over continuous translations,
encoding per-cell Bernoulli occupancy and uncertainty within
a single grid to downweight uncertain cells.

\subsection{Learning-Based Descriptors}
PointNetVLAD~\cite{uy2018pointnetvlad} aggregates PointNet features
with a NetVLAD layer for metric retrieval from raw point clouds.
OverlapNet~\cite{chen2021overlapnet} estimates pairwise overlap and
yaw offset from range images.
MinkLoc3D~\cite{komorowski2021minkloc3d} applies sparse 3D
convolutions to voxelized point clouds, achieving
strong recall with a compact descriptor.
LoGG3D-Net~\cite{vidanapathirana2022logg3d} combines local
consistency loss with a global descriptor head for geometry-aware
retrieval.
OverlapTransformer~\cite{ma2022overlaptransformer} augments a
convolutional range-image encoder with a lightweight transformer
to achieve yaw-angle invariance with sub-2\,ms inference.
BEVPlace++~\cite{luo2024bevplace++} renders point clouds into
BEV images and trains a rotation-equivariant CNN with a NetVLAD
head to produce rotation-invariant compact global descriptors with fast
nearest-neighbor retrieval.
These methods generalize well but depend on
training data and GPU inference, which limits their use
on embedded platforms.
PROBE achieves comparable accuracy through probabilistic
modeling without any learned parameters.

\begin{figure}[!t]
  \centering
  \includegraphics[width=0.95\columnwidth]{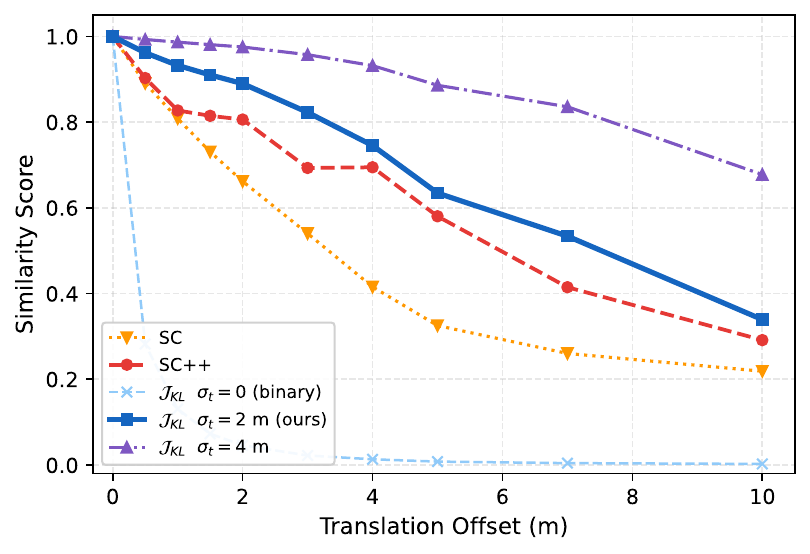}
  \caption{Translation robustness: similarity under increasing
    lateral translation on KITTI-08 (averaged over 7 frames and 3 translation directions).
    SC (no augmentation) degrades fastest; SC++ mitigates this via
    discrete augmentation, producing a characteristic plateau.
    The Bernoulli-KL Jaccard~$\mathcal{J}_{KL}$ is shown at three blur levels:
    without blur ($\sigma_t{=}0$, dashed), $\sigma_t{=}2$\,m
    (proposed), and $\sigma_t{=}4$\,m.  Analytical marginalization transforms
    a brittle binary descriptor into a smoothly robust one.}
  \label{fig:translation}
  \vspace{-2.7mm}
\end{figure}

\subsection{Local Feature Methods}
STD~\cite{yuan2023std} encodes keypoints extracted on plane
boundaries into viewpoint-invariant stable triangle descriptors
for loop closure detection within SLAM pipelines.
BTC~\cite{yuan2024btc} extends STD by combining the triangle
descriptor with a binary descriptor that encodes each keypoint's
local neighborhood, verifying matches via vertex correspondence.
While these approaches perform well when integrated with a mapping backend, 
their reliance on explicit geometric verification restricts inference speed, 
making them less suited for standalone pairwise evaluation.

\subsection{Positioning of PROBE}
{PROBE belongs to the handcrafted BEV family but replaces
deterministic binary matching with probabilistic occupancy
modeling via analytical marginalization and uncertainty-gated
scoring (Sec.~\ref{sec:method}).
Like SC and SC++, it requires no training and no geometric
verification.
For a comprehensive survey, see~\cite{zhang2024lprsurvey}.
To our knowledge, no existing method models per-cell
occupancy as a Bernoulli random variable with analytically
derived uncertainty for BEV-based place recognition.}

\section{Methodology}
\label{sec:method}

Given a 3D LiDAR point cloud
$\mathcal{P} = \{(x_i, y_i, z_i)\}_{i=1}^N$,
PROBE constructs a compact descriptor and scores pairwise similarity
in four steps (Fig.~\ref{fig:pipeline}): (i)~BEV polar grid
construction with max-height encoding,
(ii)~per-cell Bernoulli occupancy estimation via Jacobian-derived
analytical marginalization, (iii)~ring-mean retrieval key
for KD-tree pre-filtering, and (iv)~FFT-based rotation alignment
with Bernoulli-KL Jaccard scoring.

\subsection{BEV Grid Construction}
\label{sec:repr}

Following the Scan Context framework~\cite{kim2018scan}, each point
is projected into a polar BEV grid
$\bm{G} \in \R^{R \times S}$, where $R$~rings partition the
horizontal range~$[0, R_{\max}]$ and $S$~sectors partition the
azimuth~$[0, 2\pi)$.  Each cell stores the maximum height of its
constituent points; empty cells are zero.  A binary occupancy mask
\begin{equation}
  \bm{O}[r,s] =
    \begin{cases}
      1 & \text{if cell } (r,s) \text{ contains } \geq 1 \text{ point,}\\
      0 & \text{otherwise,}
    \end{cases}
  \label{eq:occ}
\end{equation}
records whether each cell is occupied.

\subsection{Per-Cell Bernoulli Occupancy Model}
\label{sec:uncertainty}

A key limitation of binary occupancy matching is that boundary cells---
those at the edge of observable structures---are
unstable: a small translational offset of the sensor origin can toggle
their occupancy state. Rather than treating all cells equally, PROBE
models each cell as a Bernoulli random variable
$\text{Bernoulli}(\mu)$ and estimates a per-cell uncertainty~$\sigma$
by analytically marginalizing over continuous translations using
the polar-grid Jacobian.  Both quantities are then exploited for
adaptive scoring in Section~\ref{sec:jccb}.

\noindent\textbf{Jacobian-derived adaptive blur.}
Consider a point at polar coordinates~$(r_0, \theta_0)$.
An isotropic Cartesian translation
$\Delta\bm{x} = (\Delta x, \Delta y)^T \sim \mathcal{N}(\bm{0}, \sigma_t^2 \bm{I})$
induces polar perturbations via the Jacobian matrix
$\bm{J} = \partial(r,\theta)/\partial(x,y)$.
The propagated polar covariance, obtained via a first-order
Taylor approximation, is
\begin{equation}
  \Sigma_{\text{polar}} \approx \bm{J} (\sigma_t^2 \bm{I}) \bm{J}^T = 
  \begin{bmatrix} 
    \sigma_t^2 & 0 \\ 
    0 & \sigma_t^2 / r_0^2 
  \end{bmatrix}.
  \label{eq:polar_cov}
\end{equation}
This yields the following decoupled perturbation distributions:
\begin{equation}
  \Delta r \sim \mathcal{N}(0, \sigma_t^2), \qquad
  \Delta \theta \sim \mathcal{N}\!\bigl(0, \sigma_t^2 / r_0^2\bigr).
  \label{eq:polar_jacobian}
\end{equation}
The angular uncertainty decreases with range: nearby cells are
more sensitive to lateral translation than distant ones, yielding
the characteristic $\sigma_\theta \propto 1/r$ relationship.
This first-order approximation is accurate when $\sigma_t \ll r_0$,
which holds for all practical ring indices beyond the first few
meters.

\noindent\textbf{Proposition 1 (Translation Marginalization).}
\textit{Let $\bm{O}[r,s] \in \{0,1\}$ be the binary occupancy of the polar grid.
The expected occupancy under isotropic Cartesian translation uncertainty
$\Delta\bm{x} \sim \mathcal{N}(\bm{0}, \sigma_t^2 \bm{I})$
is}
\begin{equation}
  \mu[r,s] = \mathbb{E}_{\Delta\bm{x}}\bigl[\bm{O}[r{+}\Delta r,\;s{+}\Delta\theta]\bigr]
  = (\bm{O} * \mathcal{G}_{\text{polar}})[r,s],
  \label{eq:marginalization}
\end{equation}
\textit{where
$\mathcal{G}_{\text{polar}}$
is a separable Gaussian kernel with widths
$\sigma_r$ and $\sigma_\theta(r)$ given by~\eqref{eq:polar_jacobian}.}

\noindent\textit{Proof sketch.}
Since $\bm{O}$ is binary, $\mathbb{E}[\bm{O}]$ reduces to the probability
that the randomly perturbed cell falls in an occupied region.
By~\eqref{eq:polar_cov}, the Cartesian perturbations map to locally
decorrelated Gaussian noise in the polar domain
($\Delta r \perp \Delta\theta$).
Thus, the 2-D expectation factorizes into two sequential 1-D
convolutions, enabling efficient separable filtering directly
on the polar grid:
\begin{align}
  \text{Step 1:} &\quad \mu[r,\cdot] \leftarrow G_{1\text{D}}\!\bigl(\bm{O}[r,\cdot];\;
    \sigma_\theta(r),\; \text{wrap}\bigr), \label{eq:angular_blur} \\
  \text{Step 2:} &\quad \mu[\cdot,s] \leftarrow G_{1\text{D}}\!\bigl(\mu[\cdot,s];\;
    \sigma_r,\; \text{constant}\bigr), \label{eq:radial_blur}
\end{align}
where $G_{1\text{D}}(\cdot;\sigma,\text{mode})$ denotes 1-D Gaussian
convolution with boundary handling,
$\sigma_\theta(r) = \sigma_{\text{eff}}(r) / (r_{\text{center}} \cdot \Delta\theta)$
is the angular kernel width in sector-cell units, and
$\sigma_r = \sigma_t / \Delta r$ is the radial kernel width in
ring-cell units, with $\Delta\theta = 2\pi/S$ and
$\Delta r = R_{\max}/R$ being the grid resolutions.
The resulting $\mu[r,s] \in [0,1]$ is exactly the marginal
occupancy probability under the assumed translation model.
\hfill$\square$

\noindent\textbf{Density-adaptive $\sigma_{\text{eff}}$.}
The Gaussian kernel assumes a continuous occupancy field;
however, LiDAR provides discrete samples whose density varies with
range and sensor resolution.  For the angular blur~\eqref{eq:angular_blur},
we scale the effective kernel bandwidth by the local occupancy density:
\begin{equation}
  \sigma_{\text{eff}}(r) = \sigma_t \cdot \sqrt{\rho(r)},
  \label{eq:density_adaptive}
\end{equation}
where $\rho(r) = \frac{1}{S}\sum_{s} \bm{O}[r,s]$ is the per-ring occupancy
rate.  The $\sqrt{\rho}$ factor scales the blur proportionally to
the observed density, preventing overconfident smoothing when
few cells are populated.
Dense rings ($\rho \!\approx\! 1$) receive full blur,
preserving the theoretical uncertainty model.  Sparse rings
($\rho \!\ll\! 1$) receive reduced blur, preventing over-smoothing
that would obscure local structures.
The radial blur~\eqref{eq:radial_blur} uses the theoretical
constant $\sigma_r = \sigma_t / \Delta r$ uniformly, since
the Jacobian-derived radial uncertainty~\eqref{eq:polar_jacobian}
is independent of position.

The per-cell Bernoulli uncertainty $\sigma$ is then derived directly
from the blurred occupancy~$\mu$.
Since the original mask~$\bm{O}$ is binary ($\bm{O}^2 = \bm{O}$),
the variance simplifies to:
\begin{equation}
  \sigma[r,s] = \sqrt{\mu[r,s] - (\mu[r,s])^2} = \sqrt{\mu[r,s](1 - \mu[r,s])}\,,
  \label{eq:sigma_bernoulli}
\end{equation}
attaining a maximum of $\sigma = 0.5$ for maximally uncertain cells ($\mu = 0.5$)
and vanishing for confident cells ($\mu \approx 0$ or $\mu \approx 1$).

\noindent\textbf{Interpretation.}
Cells deep within a structure's footprint retain $\mu \approx 1$
after blurring, yielding $\sigma \approx 0$: the blur merely
reinforces the occupied region.  Boundary cells where the blur
spreads occupancy across the occupied/empty edge produce intermediate
$\mu$ and correspondingly high~$\sigma$, signaling low confidence.
The density-adaptive scaling ensures that this mechanism scales
down on sparse sensors (e.g., 32-beam LiDAR) by reducing
blur where the sampling density is insufficient to support the
assumed convolution.
Fig.~\ref{fig:translation} illustrates the resulting
translation robustness.

\subsection{Ring-Mean Retrieval Key}
\label{sec:retrieval}

{For large-scale place recognition, exhaustive pairwise scoring is
prohibitive. Following SC~\cite{kim2018scan}, PROBE constructs a compact,
rotation-invariant \emph{retrieval key} for KD-tree nearest-neighbor pre-filtering.}

\begin{figure}[!t]
  \centering
  \includegraphics[width=0.95\columnwidth]{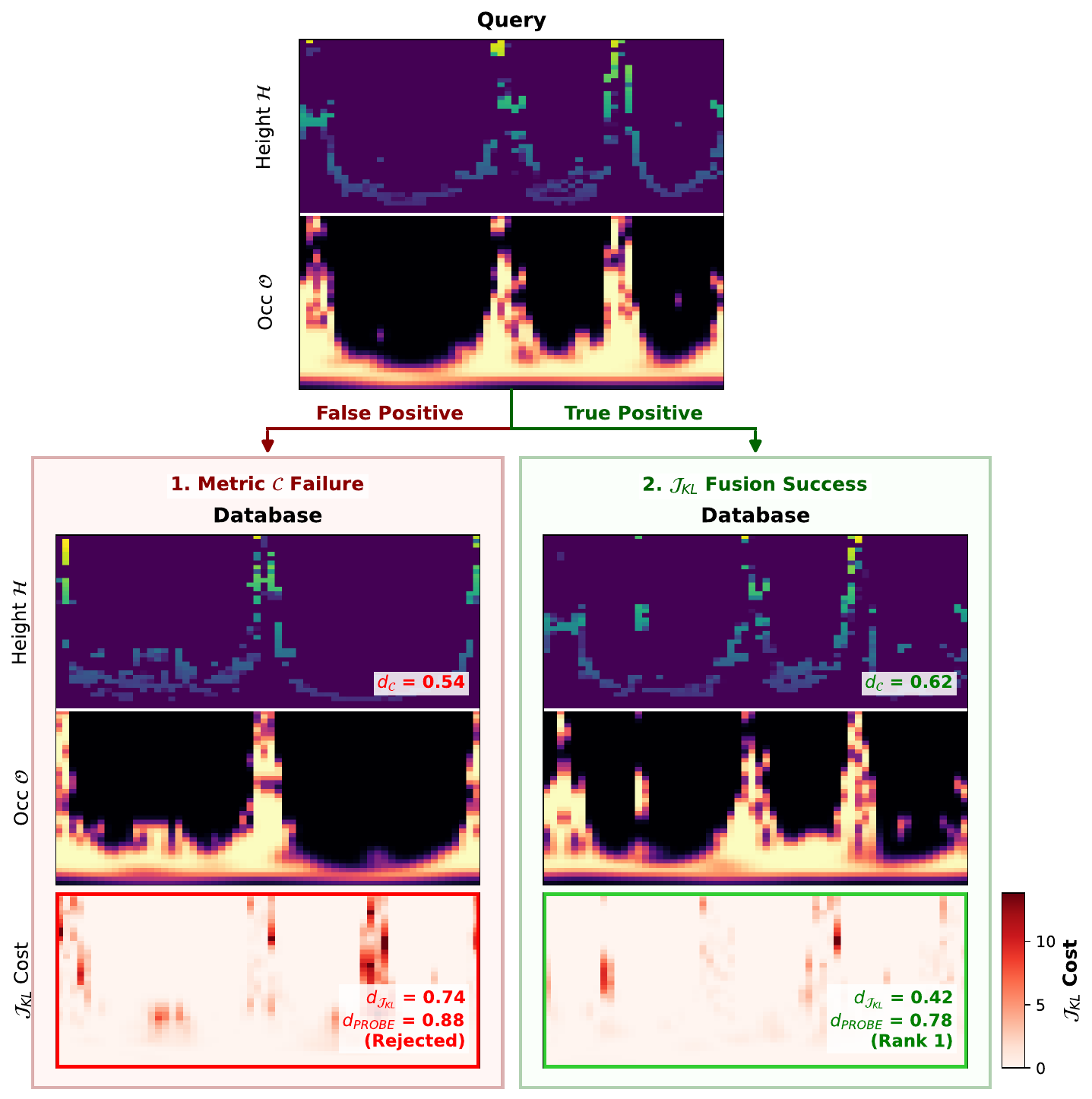}
  \caption{Discriminative capability of the
    $\mathcal{J}_{KL}$ mechanism.
    (Left)~A False Positive from a structurally different location
    achieves a low height distance ($d_{\mathcal{C}}\!=\!0.54$)
    because $\mathcal{C}$ relies exclusively on the sparse max-height
    envelope, which can spuriously align between structurally
    different scenes.
    However, the per-cell KL cost map evaluates dense occupancy
    probabilities, exposing spatial mismatches (bright regions),
    and the fused PROBE metric correctly rejects it.
    (Right)~The True Positive yields a uniformly low-cost map
    and is correctly retrieved as Rank~1.}
  \label{fig:jkl_mechanism}
  \vspace{-2.7mm}
\end{figure}

{Since the azimuthal mean over all sectors in a ring is
invariant to circular shifts, PROBE concatenates two complementary
ring-mean vectors into a $2R$-dimensional key:
\begin{equation}
  \bm{k} = \bigl[\bar{\bm{G}}\;\|\;\bar{\bm{\mu}}\bigr]
  \in \mathbb{R}^{2R},
  \label{eq:ring_key}
\end{equation}
where $\bar{\bm{G}}[r] = \frac{1}{S}\sum_{s} \bm{G}[r,s]$ captures the
height profile, and $\bar{\bm{\mu}}[r] = \frac{1}{S}\sum_{s} \bm{\mu}[r,s]$
captures the occupancy density, inheriting translation robustness
from the analytical marginalization.
Top-$K$ candidates retrieved via Euclidean distance are subsequently
re-ranked using the full pairwise score~\eqref{eq:distance}.}

\subsection{Rotation Alignment and Pairwise Scoring}
\label{sec:jccb}

For each retrieved candidate pair, PROBE first resolves the heading
ambiguity via FFT-accelerated circular cross-correlation over the
max-height grid, then fuses two complementary probabilistic cues
into a final similarity score.

\subsubsection{FFT-Based Rotation Alignment}
\label{sec:rotation}
The normalized circular cross-correlation at azimuthal shift~$\delta$ is
\begin{equation}
  \text{CC}[\delta] = \frac{
    \displaystyle\sum_{r=0}^{R-1}\sum_{s=0}^{S-1}
    G_m[r, s] \; G_q[r, (s{+}\delta) \bmod S]
  }{
    \norm{\bm{G}_m}_F \cdot \norm{\bm{G}_q}_F
  }\,,
  \label{eq:cc_global}
\end{equation}
computed in $\mathcal{O}(R \cdot S + S \log S)$ via the convolution
theorem using pre-computed row-wise FFTs.
The optimal rotation estimate is
\begin{equation}
  \delta^* = \argmax_{\delta \in \{0, \ldots, S-1\}} \text{CC}[\delta].
  \label{eq:map}
\end{equation}

\subsubsection{Bernoulli-KL Jaccard ($\mathcal{J}_{KL}$)}
Classical binary Jaccard treats all cells in the union mask
identically, which penalizes boundary mismatches as heavily as
interior mismatches.  We replace it with a {\em Bernoulli-KL Jaccard}
that models each cell probabilistically and downweights unreliable
cells.

Let $\mu_m, \mu_q, \sigma_m, \sigma_q$ denote the Bernoulli mean and
uncertainty maps of the map and rotationally-aligned query,
respectively.  Define the {\em soft union set}
$\mathcal{U} = \{(r,s) : \mu_m[r,s] + \mu_q[r,s] > {\varepsilon_U}\}$
(${\varepsilon_U} = 10^{-3}$), where the blurred occupancy
probabilities~$\mu$ define the support region rather than
the binary mask~$\bm{O}$.  This ensures that boundary cells
whose Gaussian-spread occupancy extends beyond the original
footprint participate in the divergence computation, consistent
with the continuous marginalization framework.

\begin{table}[!t]
  \centering
  \caption{Computational Complexity and Empirical Runtime.
    $N$: points, $R$: rings, $S$: sectors, $H$: height bins,
    $K$: column search window ($K \leq S$){, $K'$: number of lateral offsets}.
    {Runtime measured on a single-threaded Intel Core i7
    using KITTI~00.}}
  \label{tab:complexity}
  \footnotesize
  \setlength{\tabcolsep}{2.5pt}
  \begin{tabular}{l cc cc}
    \toprule
    & \multicolumn{2}{c}{\textbf{Complexity (Big-$\mathcal{O}$)}}
    & \multicolumn{2}{c}{\textbf{Runtime (ms)}} \\
    \cmidrule(lr){2-3} \cmidrule(lr){4-5}
    \textbf{Method} & \textbf{Build} & \textbf{Match}
    & {\textbf{Extr.}} & {\textbf{Match}} \\
    \midrule
    SC~\cite{kim2018scan}   & $\mathcal{O}(N)$ & $\mathcal{O}(K{\cdot}R{\cdot}S)$
      & {\textbf{36.29}} & {1.518} \\
    SC++~\cite{kim2022sc++} & $\mathcal{O}(K'{\cdot}N)$ & $\mathcal{O}(K{\cdot}R{\cdot}S)$
      & {38.14} & {1.522} \\
    SOLiD~\cite{kim2024narrowing} & $\mathcal{O}(N{+}R{\cdot}H)$ & $\mathcal{O}(R)$
      & {38.20} & {\textbf{0.002}} \\
    \textbf{PROBE} & $\mathcal{O}(N{+}R{\cdot}S\!\log\!S)$ & $\mathcal{O}(R{\cdot}S{+}S\!\log\!S)$
      & {\underline{36.72}} & {\underline{0.085}} \\
    \bottomrule
  \end{tabular}
  \vspace{-2.7mm}
\end{table}

\noindent\textbf{Step 1: Uncertainty-proportional shrinkage.}
Rather than using the raw marginal probability~$\mu$ directly,
we shrink each cell's occupancy toward the uninformative prior
$p = 0.5$ in proportion to its Bernoulli uncertainty~$\sigma$:
\begin{equation}
  p_m[r,s] = \mu_m[r,s] \cdot (1 - \sigma_m[r,s]) + 0.5 \cdot \sigma_m[r,s],
  \label{eq:smooth_p}
\end{equation}
{and identically for the query, $p_q[r,s]$, using
$(\mu_q, \sigma_q)$.  Both are}
strictly bounded within $[\varepsilon_B,\; 1 - \varepsilon_B]$ for numerical stability
($\varepsilon_B = 10^{-6}$).
When $\sigma \to 0$ (confident cell), $p \to \mu$ and
the observation dominates;
when $\sigma \to 0.5$ (maximally uncertain boundary cell),
$p \to 0.5$ and the cell is effectively neutralized,
contributing zero KL divergence in subsequent steps.

\noindent\textbf{Step 2: Symmetric KL divergence.}
For each cell $(r,s) \in \mathcal{U}$, the symmetric KL divergence
between the smoothed Bernoulli distributions is
\begin{equation}
  D_{KL}[r,s] = \tfrac{1}{2}\bigl(
    \text{KL}(p_m \| p_q) + \text{KL}(p_q \| p_m)\bigr),
  \label{eq:sym_kl}
\end{equation}
where $\text{KL}(p \| q) = p \ln(p/q) + (1{-}p)\ln((1{-}p)/(1{-}q))$.

\noindent\textbf{Bernoulli-KL Jaccard.}
The Bernoulli-KL Jaccard\footnote{We retain the term ``Jaccard''
because, like the binary Jaccard index, $\mathcal{J}_{KL}$ is
defined over the union of occupied cells and returns $1$ for
identical inputs.} aggregates the per-cell divergence via the
arithmetic mean over the soft union~$\mathcal{U}$:
\begin{equation}
  \mathcal{J}_{KL}(\delta^*) = \exp\!\left(
    -\frac{1}{|\mathcal{U}|}
    \sum_{(r,s) \in \mathcal{U}} D_{KL}[r,s]\right) \in (0, 1].
  \label{eq:kl_jaccard}
\end{equation}
Explicit confidence weighting is unnecessary: the shrinkage in
Step~1 already drives uncertain cells toward $p{=}0.5$, where
$D_{KL} \to 0$, so they contribute negligible divergence.

\noindent\textbf{Analysis.}
When both scans agree ($p_m = p_q$), $D_{KL} = 0$ and
$\mathcal{J}_{KL} = 1$.
When they disagree, the shrinkage attenuates unreliable cells:
Bayesian shrinkage (Step~1) drives uncertain occupancies toward
the prior, reducing their KL contribution to near zero.
Stable interior mismatches, where both $\sigma$ values are near
zero, dominate the mean KL, producing low $\mathcal{J}_{KL}$
and correct rejection.

\noindent\textbf{Remark (Component Interaction).}
The three scoring components form a cascaded uncertainty pipeline:
(i)~Gaussian blur converts binary occupancy into continuous
probabilities~$\mu$, producing per-cell Bernoulli uncertainty~$\sigma$
(Sec.~\ref{sec:uncertainty});
(ii)~shrinkage drives uncertain cells ($\sigma \!\to\! 0.5$) toward the
uninformative prior $p{=}0.5$, neutralizing their KL contribution
(Step~1);
(iii)~symmetric KL aggregation over the soft union~$\mathcal{U}$
measures structural disagreement only from confident cells (Step~2).
This cascade ensures that the final score~$\mathcal{J}_{KL}$ is
dominated by geometrically stable regions, not volatile boundaries
(Fig.~\ref{fig:jkl_mechanism}).

\subsubsection{Cosine Similarity ($\mathcal{C}$)}
The cosine similarity at the aligned rotation~$\delta^*$ is
\begin{equation}
  \mathcal{C}(\delta^*) = \text{CC}[\delta^*]\,,
  \label{eq:cosine}
\end{equation}
i.e., the normalized cross-correlation value at the estimated
rotation offset~\eqref{eq:map}.

\subsubsection{Final Distance: Log-Linear Evidence Fusion}
The PROBE similarity combines two complementary cues:
\begin{equation}
  S_{\text{PROBE}} = \mathcal{J}_{KL}(\delta^*) \cdot
  \mathcal{C}(\delta^*),
  \label{eq:distance}
\end{equation}
with distance $d(m,q) = 1 - S_{\text{PROBE}}$.

\begin{table}[!t]
  \centering
  \caption{PROBE Hyper-Parameters (Fixed Across All Experiments)}
  \label{tab:params}
  \begin{tabular}{llc}
    \toprule
    \textbf{Symbol} & \textbf{Description} & \textbf{Value} \\
    \midrule
    $R$ & Number of distance rings & 40 \\
    $S$ & Number of azimuth sectors & 60 \\
    $R_{\max}$ & Maximum range~[\si{\metre}] & 80 \\
    $\sigma_t$ & Translation uncertainty~[\si{\metre}] & 2.0 \\
    $\varepsilon_B$ & Bernoulli clamp & $10^{-6}$ \\
    $v$ & Voxel grid size~[\si{\metre}] & 0.5 \\
    \bottomrule
  \end{tabular}
  \vspace{-2.7mm}
\end{table}

\noindent\textbf{Probabilistic interpretation.}
Defining $\ell_1 = \log \mathcal{J}_{KL}$ and
$\ell_2 = \log \mathcal{C}$ as log-likelihood contributions
from occupancy and height evidence respectively, the
multiplicative score becomes
\begin{equation}
  \log S_{\text{PROBE}} = \ell_1 + \ell_2,
  \label{eq:loglinear}
\end{equation}
which is a \emph{log-linear model}~\cite{bishop2006pattern}
that aggregates conditionally independent evidence sources.
Under the assumption that occupancy agreement and height
correlation are independent conditioned on the event
``same place,'' the product form yields the joint likelihood
$P(\text{data} \mid \text{same place}) \propto
\mathcal{J}_{KL} \cdot \mathcal{C}$.
The multiplicative structure ensures that \emph{both} cues
must agree for a high score: a single sufficiently negative
cue (low $\mathcal{J}_{KL}$ or low $\mathcal{C}$)
suffices for rejection, consistent with the
veto property of likelihood-based fusion.
{
As illustrated in Fig.~\ref{fig:jkl_mechanism}, this complementarity
allows the fused metric to successfully reject perceptually
aliased false positives that share a numerically similar height
profile but differ in structural occupancy.}

\subsection{Computational Complexity}
\label{sec:complexity}

Table~\ref{tab:complexity} compares computational costs
for descriptor construction and pairwise matching.
All methods share the $\mathcal{O}(N)$ grid-building pass;
PROBE adds an azimuthal FFT and separable Gaussian blur
per ring, both negligible relative to the point-cloud pass.
For pairwise matching, SC exhaustively evaluates $K$ column
shifts at $\mathcal{O}(K \cdot R \cdot S)$, whereas PROBE
resolves rotation via FFT cross-correlation in
$\mathcal{O}(R \cdot S + S \log S)$ using pre-computed spectra,
yielding a log-factor improvement.
SOLiD achieves the lowest per-pair cost via compact vector
distance ($\mathcal{O}(R)$).
{
As empirically validated in Table~\ref{tab:complexity},
PROBE's extraction time remains comparable to standard baselines,
and its FFT-based matching is substantially faster
than SC's exhaustive column shifting.
}
\section{Experiments}
\label{sec:experiments}

\begin{table*}[!t]
  \centering
  \caption{Single-Session Place Recognition (AUC, Online Mode, $d_{\text{gt}}{=}\SI{10}{\metre}$).
    V-64/32/16: Velodyne HDL-64/32/16; O-128: Ouster OS2-128;
    CU: ComplexUrban.
    Best in \textbf{bold}, second-best \underline{underlined}.
    $\dagger$: supervised (trained on KITTI).}
  \label{tab:single_auc}
  \footnotesize
  \setlength{\tabcolsep}{3.0pt}
  \begin{tabular}{l cccc cccc cccc cccc c}
    \toprule
    & \multicolumn{4}{c}{\textbf{KITTI (V-64)}}
    & \multicolumn{4}{c}{\textbf{HeLiPR (O-128)}}
    & \multicolumn{4}{c}{\textbf{NCLT (V-32)}}
    & \multicolumn{4}{c}{\textbf{CU (V-16)}}
    & \\
    \cmidrule(lr){2-5} \cmidrule(lr){6-9}
    \cmidrule(lr){10-13} \cmidrule(lr){14-17}
    \textbf{Method}
    & 00 & 02 & 05 & 08
    & k05 & k06 & r05 & r06
    & 0526 & 0820 & 0928 & 0405
    & 00 & 01 & 02 & 04
    & \textbf{Avg} \\
    \midrule
    M2DP~\cite{he2016m2dp}
    & .889 & .731 & .710 & .005
    & .525 & .015 & .846 & .801
    & .356 & .077 & .224 & .193
    & \underline{.151} & .672 & .218 & .550
    & .435 \\
    SC~\cite{kim2018scan}
    & .901 & .681 & .738 & .644
    & .887 & .923 & .830 & .648
    & .677 & .524 & .607 & .605
    & .045 & .388 & .335 & .476
    & .619 \\
    LiDAR-Iris~\cite{wang2020lidar}
    & .879 & .764 & .726 & .614
    & .885 & .918 & .876 & .624
    & .725 & .518 & .606 & .648
    & .037 & .306 & .352 & .422
    & .619 \\
    SC++~\cite{kim2022sc++}
    & .905 & .830 & .760 & .785
    & \underline{.957} & \underline{.982} & .882 & .819
    & .761 & .628 & .692 & .719
    & .120 & .619 & .354 & .504
    & .707 \\
    RING++~\cite{lu2023deepring}
    & .826 & .755 & \textbf{.859} & .704
    & .911 & .933 & \textbf{.981} & \textbf{.968}
    & .776 & .686 & .625 & .660
    & \textbf{.818} & \textbf{.905} & .296 & \textbf{.815}
    & \textbf{.782} \\
    SOLiD~\cite{kim2024narrowing}
    & .892 & .674 & .710 & .752
    & .706 & .618 & .734 & .563
    & .492 & .344 & .491 & .591
    & .008 & .136 & .142 & .265
    & .507 \\
    BEVPlace++$^\dagger$~\cite{luo2024bevplace++}
    & \underline{.911} & \textbf{.901} & \underline{.775} & \textbf{.948}
    & \textbf{.983} & \textbf{.989} & .854 & \underline{.845}
    & \underline{.847} & \underline{.758} & \textbf{.815} & \underline{.798}
    & .058 & .526 & \underline{.377} & .568
    & \underline{.747} \\
    \midrule
    \textbf{PROBE (Ours)}
    & \textbf{.912} & \underline{.782} & .754 & \underline{.794}
    & .951 & .958 & \underline{.901} & .761
    & \textbf{.882} & \textbf{.776} & \underline{.776} & \textbf{.817}
    & .123 & \underline{.712} & \textbf{.390} & \underline{.616}
    & .744 \\
    \bottomrule
  \end{tabular}
  \vspace{-2.7mm}
\end{table*}

\subsection{Setup}

\begin{figure}[!t]
  \centering
  \includegraphics[width=0.95\columnwidth]{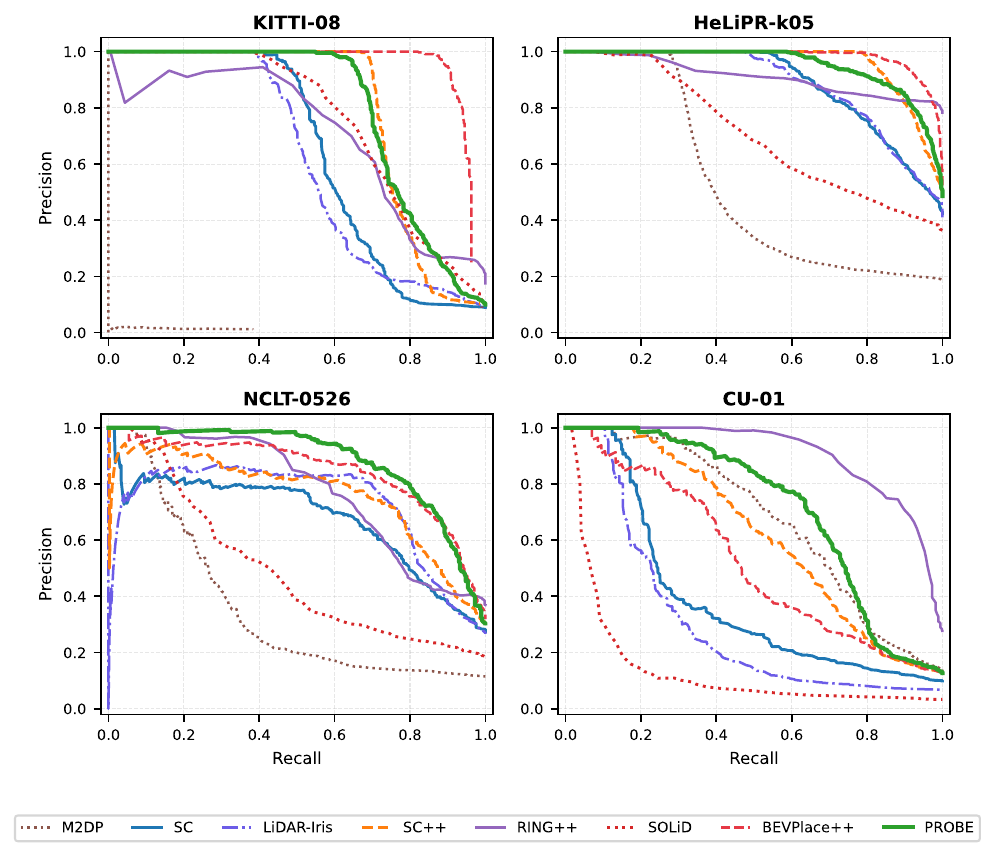}
  \caption{Single-session Precision-Recall curves
    (one representative sequence per dataset).}
  \label{fig:pr_single}
  \vspace{-2.7mm}
\end{figure}

\noindent\textbf{Datasets.}
We evaluate on four datasets spanning four LiDAR types:
\textbf{KITTI}~\cite{geiger2012kitti} (Velodyne HDL-64E, sequences
00/02/05/08),
\textbf{HeLiPR}~\cite{jung2024helipr} (Ouster OS2-128, campus and
riverside),
\textbf{NCLT}~\cite{carlevaris2016nclt} (Velodyne HDL-32E, four campus sessions), and
\textbf{ComplexUrban}~\cite{jeong2019complexurban} (Velodyne VLP-16, dense urban).
In total, we report 16 single-session and 6 multi-session sequence
pairs (22 configurations).

\noindent\textbf{Evaluation.}
Single-session matching uses \emph{online} mode:
each query retrieves from past frames with a
$\SI{25}{\metre}$ trajectory-distance exclusion zone.
Multi-session uses the full database.
Ground-truth positives are pairs within~$\SI{10}{\metre}$
for all datasets (twice the $\SI{5}{\metre}$ database
sampling interval).

\noindent\textbf{Baselines.}
M2DP~\cite{he2016m2dp},
Scan Context~(SC)~\cite{kim2018scan},
LiDAR-Iris~\cite{wang2020lidar},
SC++~\cite{kim2022sc++},
RING++~\cite{lu2023deepring},
SOLiD~\cite{kim2024narrowing},
and BEVPlace++~\cite{luo2024bevplace++} (supervised$^\dagger$; trained on KITTI).

\noindent\textbf{Metrics.}
Precision-Recall~(PR) curves, \textbf{AUC} (area under the PR curve),
\textbf{R@1} (recall at rank~1), and
\textbf{F1$_{\max}$} (maximum F1 score).
Tables report AUC; R@1 and F1$_{\max}$ follow consistent
trends and are omitted for conciseness.

\noindent\textbf{Parameters.}
PROBE uses fixed hyper-parameters across all experiments
(Table~\ref{tab:params}).

\subsection{Single-Session Results}

Table~\ref{tab:single_auc} reports the AUC for 16 single-session
sequences across four diverse sensor configurations
(Fig.~\ref{fig:pr_single} shows representative PR curves).
Overall, RING++ achieves the highest average AUC,
followed closely by BEVPlace++ and PROBE.
PROBE ranks third on average, competitive on {32--128-beam} sensors
but limited on the sparse 16-beam configuration.

\noindent\textbf{KITTI (V-64).}
On the dense 64-beam KITTI dataset, the supervised BEVPlace++ yields
the highest dataset average, benefiting from being trained
directly on this domain.
Among the handcrafted methods, PROBE achieves competitive results,
leading on seq.~00.
PROBE's continuous formulation does not sacrifice
discriminability even in dense environments where binary occupancy
is inherently stable.

\noindent\textbf{HeLiPR (O-128).}
The HeLiPR dataset features the ultra-dense Ouster OS2-128 sensor.
Here, RING++ achieves the strongest performance on the riverside
sequences, using its translation-invariant
Radon-Fourier features that excel under dense spatial sampling.
Conversely, BEVPlace++ leads the KAIST sequences.
PROBE maintains competitive performance throughout,
confirming that the probabilistic blur does not degrade
discriminability on high-density grids.

\noindent\textbf{NCLT (V-32).}
PROBE achieves the highest dataset average on the 32-beam NCLT
dataset, outperforming all baselines including the supervised
BEVPlace++, for which NCLT constitutes an unseen domain.
These results indicate that the
Jacobian-derived $\sigma$-gating is effective on sensors
{with} sufficient angular density for statistical modeling,
yet sparse enough that boundary cells suffer from thresholding
artifacts under binary occupancy.

\noindent\textbf{ComplexUrban (V-16).}
The VLP-16 sensor produces inherently sparse and fragmented BEV grids,
posing a {considerable} challenge for geometry-based descriptors.
RING++ maintains robust performance due to the global
integration properties of the Radon transform.
All direct BEV matching methods, including the supervised BEVPlace++
and PROBE, exhibit limited performance on this dataset.
Under such extreme sparsity, the continuous KL divergence estimates become
noisy, as too few cells are occupied to form a reliable
occupancy distribution.

\begin{table}[!t]
  \centering
  \caption{Multi-Session Place Recognition (AUC, $d_{\text{gt}}{=}\SI{10}{\metre}$).
    6 cross-session pairs across two datasets.
    Best in \textbf{bold}, second-best \underline{underlined}.
    $\dagger$: supervised (trained on KITTI).}
  \label{tab:inter_auc}
  \footnotesize
  \setlength{\tabcolsep}{4pt}
  \begin{tabular}{l ccc ccc c}
    \toprule
    & \multicolumn{3}{c}{\textbf{HeLiPR (O-128)}}
    & \multicolumn{3}{c}{\textbf{NCLT (V-32)}}
    & \\
    \cmidrule(lr){2-4} \cmidrule(lr){5-7}
    \textbf{Method}
    & \rotatebox{70}{k5$\to$6}
    & \rotatebox{70}{r5$\to$6}
    & \rotatebox{70}{r6$\to$5}
    & \rotatebox{70}{26$\to$20}
    & \rotatebox{70}{26$\to$28}
    & \rotatebox{70}{26$\to$05}
    & \textbf{Avg} \\
    \midrule
    M2DP~\cite{he2016m2dp}
    & .869 & .824 & .806
    & .808 & .797 & .214
    & .720 \\
    SC~\cite{kim2018scan}
    & .983 & .861 & .876
    & .927 & .914 & .648
    & .868 \\
    LiDAR-Iris~\cite{wang2020lidar}
    & \textbf{.993} & .931 & \textbf{.963}
    & .932 & .942 & \underline{.815}
    & .929 \\
    SC++~\cite{kim2022sc++}
    & .980 & .860 & .862
    & .931 & .926 & .603
    & .860 \\
    RING++~\cite{lu2023deepring}
    & .837 & .827 & .934
    & .822 & .779 & .291
    & .748 \\
    SOLiD~\cite{kim2024narrowing}
    & .683 & .551 & .578
    & .788 & .705 & .228
    & .589 \\
    BEVPlace++$^\dagger$~\cite{luo2024bevplace++}
    & .973 & \textbf{.945} & .920
    & \textbf{.981} & \textbf{.977} & \textbf{.851}
    & \textbf{.941} \\
    \midrule
    \textbf{PROBE (Ours)}
    & \underline{.986} & \underline{.941} & \underline{.940}
    & \underline{.979} & \textbf{.977} & .785
    & \underline{.935} \\
    \bottomrule
  \end{tabular}
  \vspace{-2.7mm}
\end{table}

\subsection{Multi-Session Results}

\begin{figure}[!t]
  \centering
  \includegraphics[width=0.97\columnwidth]{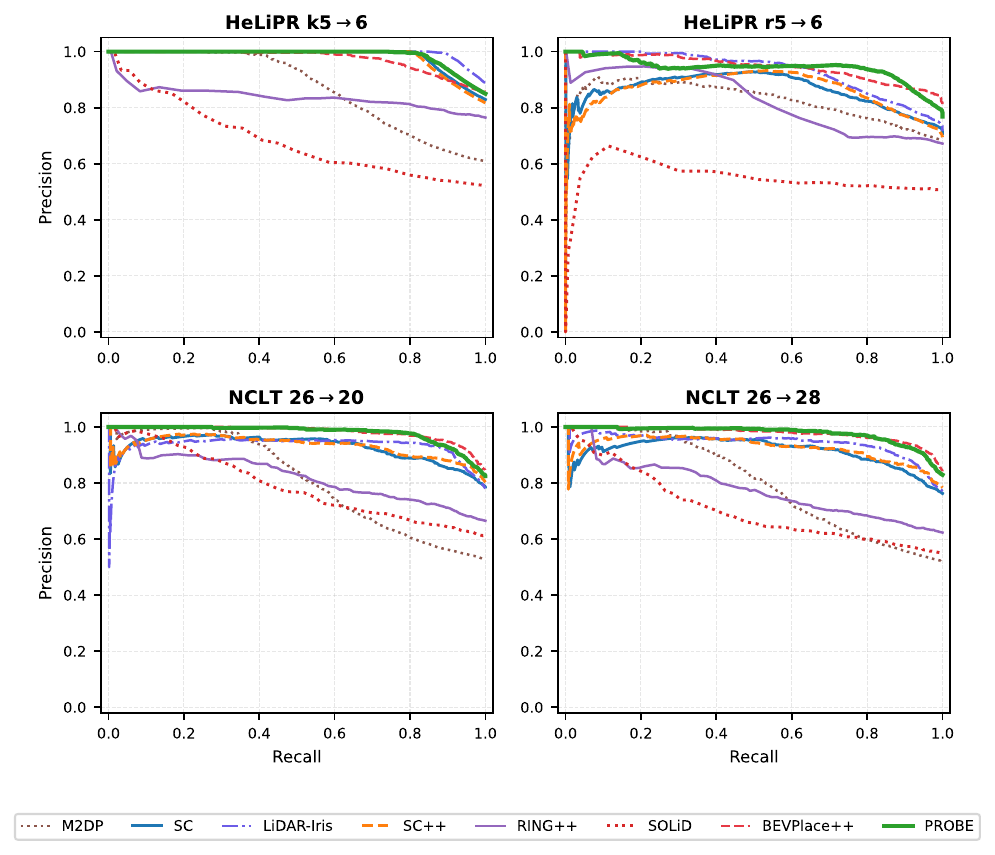}
  \caption{Multi-session Precision-Recall curves
    (two representative pairs per dataset, four total).}
  \label{fig:pr_multi}
  \vspace{-2.7mm}
\end{figure}

Table~\ref{tab:inter_auc} reports AUC for 6 multi-session pairs
across HeLiPR and NCLT; Fig.~\ref{fig:pr_multi} shows
representative PR curves.
Multi-session evaluation is more challenging due to
spatiotemporal variations and differing traversal trajectories.

The supervised BEVPlace++ achieves the highest average AUC.
PROBE follows closely, achieving the best results
among handcrafted methods and outperforming LiDAR-Iris.
RING++, which dominates single-session evaluation,
ranks sixth in the multi-session average.
While Radon-Fourier features provide translation invariance
for locally dense revisits, they lose discriminability when
global map and query trajectories diverge across sessions.

\noindent\textbf{HeLiPR.}
LiDAR-Iris achieves the highest scores on specific sequences.
PROBE, however, demonstrates the most consistent robustness,
ranking second across all three pairs despite
substantial spatiotemporal structural changes.

\noindent\textbf{NCLT.}
Despite not being trained on NCLT, the learned features of BEVPlace++
generalize {to this unseen domain}, achieving the highest AUCs.
PROBE closely follows the supervised baseline, BEVPlace++,
matching it on 26$\to$28.
The analytical $\sigma$-gating accommodates the occupancy variations
that otherwise degrade binary matching.

\subsection{Ablation and Robustness Analysis}

\begin{table}[!t]
  \centering
  \caption{Ablation Summary (AUC, $d_{\text{gt}}{=}\SI{10}{\metre}$).
    Best in each group in \textbf{bold},
    second-best \underline{underlined}.}
  \label{tab:ablation}
  \begin{tabular}{l l c c}
    \toprule
    \textbf{Group} & \textbf{Variant}
    & \textbf{Single} & \textbf{Multi} \\
    \midrule
    \multirow{3}{*}{\rotatebox{90}{\scriptsize Score}}
      & (C1) $\mathcal{C}$ only  & \underline{.711} & \underline{.915} \\
      & (C2) $\mathcal{J}_{KL}$ only  & .691 & .902 \\
      & \textbf{(C3) $\mathcal{J}_{KL} \cdot \mathcal{C}$ (ours)} & \textbf{.744} & \textbf{.935} \\
    \midrule
    \multirow{3}{*}{\rotatebox{90}{\scriptsize Blur}}
      & $\sigma_t = 0$ (binary)  & .679 & .903 \\
      & \textbf{$\sigma_t = 2$\,m (ours)}  & \textbf{.744} & \textbf{.935} \\
      & $\sigma_t = 4$\,m  & \underline{.739} & \underline{.931} \\
    \bottomrule
  \end{tabular}
  \vspace{-2.7mm}
\end{table}

\begin{figure}[!t]
  \centering
  \includegraphics[width=0.90\columnwidth]{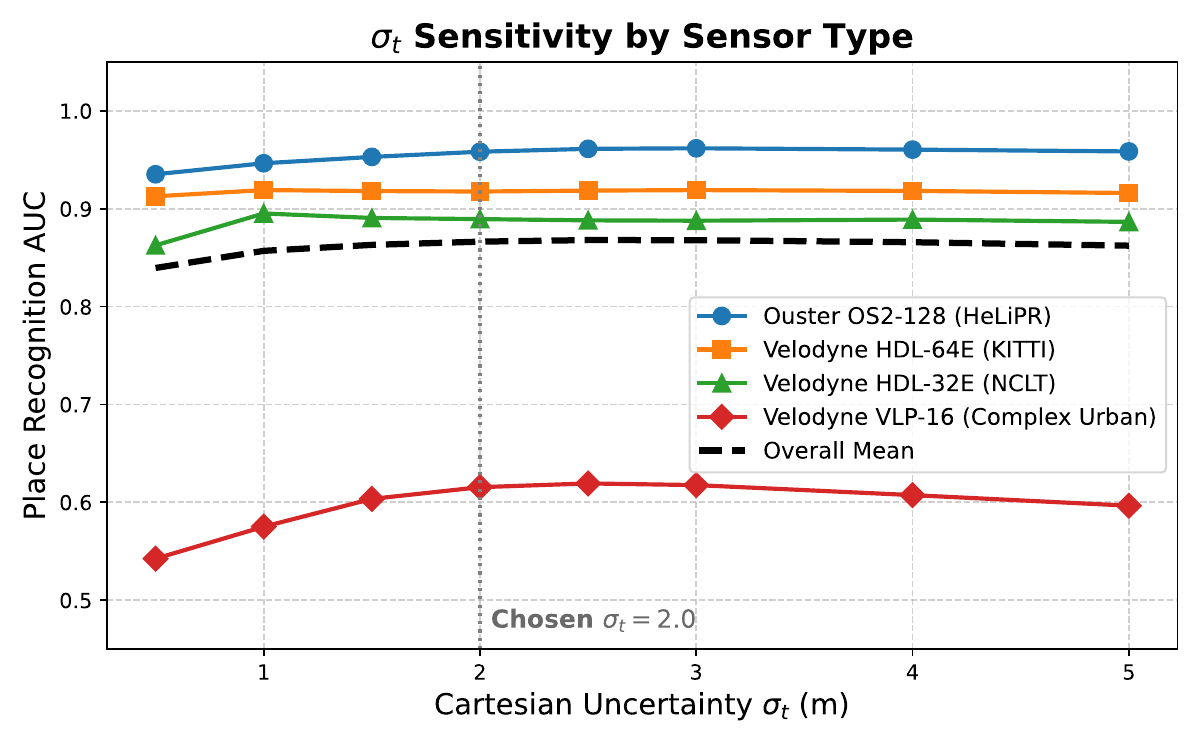}
  \caption{Sensitivity of $\sigma_t$ across four
    LiDAR sensor types.  The AUC remains stable between
    $1.5$--$3.0$\,m, confirming $\sigma_t{=}2.0$\,m
    as a robust default.}
  \label{fig:sigma_sensitivity}
  \vspace{-2.7mm}
\end{figure}

{\noindent\textbf{Score components and $\sigma_t$ sensitivity.}}
{Table~\ref{tab:ablation} summarizes ablations averaged across all
22 sequence configurations.
The multiplicative fusion \textbf{(C3)} substantially outperforms
individual cues, confirming that occupancy agreement and height
alignment provide complementary evidence.}
{Disabling the Jacobian blur ($\sigma_t{=}0$) reverts the system to
binary matching and degrades the AUC, validating the contribution
of the Bernoulli layer.
A finer-grained sweep over eight values ($0.5$--$5.0$\,m)
on four LiDAR configurations
(Fig.~\ref{fig:sigma_sensitivity}) reveals a broad plateau
between $\sigma_t{=}1.5$\,m and $3.0$\,m across all sensors,
with less than $0.5\%$ variance.
Denser sensors tend toward larger values
($\sim 3.0$\,m), while sparse sensors peak near $2.5$\,m;
the default $\sigma_t{=}2.0$\,m remains within the plateau,
supporting its sensor-agnostic robustness.}

{
\noindent\textbf{Robustness under restricted FOVs.}
We simulated restricted FOVs under symmetric and asymmetric
conditions (Fig.~\ref{fig:fov_ablation}).
Under symmetric clipping, forward loops degrade gracefully,
while reverse loops drop to $0\%$ recall as expected.
Under asymmetric restriction (narrow query vs.~$360^\circ$ DB),
PROBE's $\sigma$-gating treats the unobserved sectors as
pure uncertainty ($\mu{=}0.5$, $\sigma{=}0.5$), allowing partial-geometry alignment.
With brute-force matching (PROBE-BF, bypassing retrieval keys
distorted by missing sectors), it recovers $0.79$~R@1
at $180^\circ$ on KITTI~08, substantially outperforming all baselines.

\begin{figure}[!t]
  \centering
  \includegraphics[width=0.97\columnwidth]{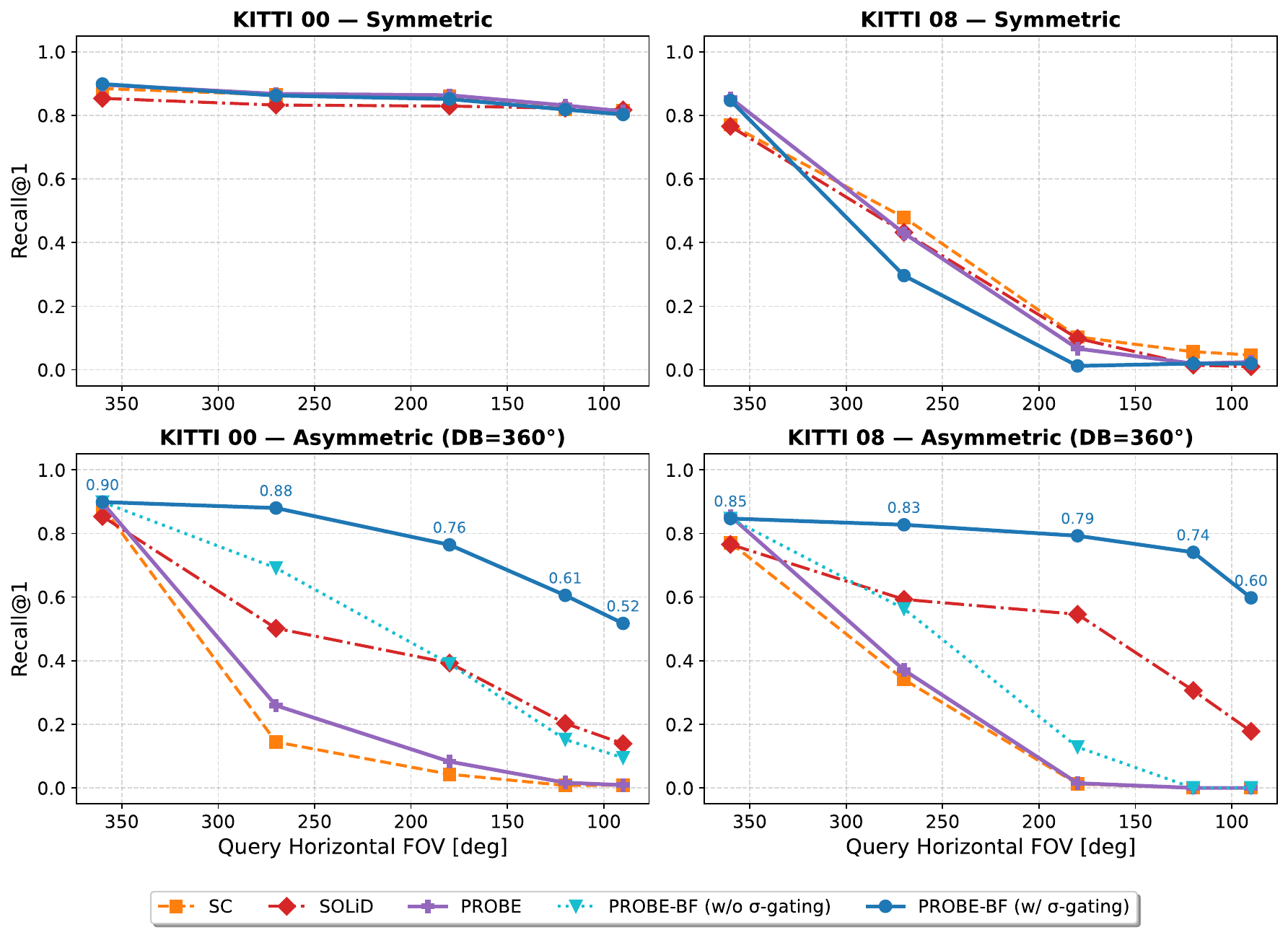}
  \caption{Place recognition performance under restricted FOVs.
    \textbf{Top (Symmetric):} Both database and query are clipped identically.
    \textbf{Bottom (Asymmetric):} A narrow-FOV query against a $360^\circ$ DB.
    KITTI~08 highlights the geometric collapse in symmetric reverse loops
    and PROBE-BF's robust recovery via $\sigma$-gating
    in asymmetric scenarios.}
  \label{fig:fov_ablation}
  \vspace{-2.7mm}
\end{figure}
}

\section{Discussion and Conclusion}
\label{sec:conclusion}

We presented PROBE, a probabilistic BEV descriptor that
addresses viewpoint sensitivity in LiDAR place recognition.
By replacing heuristic binary occupancy with a per-cell Bernoulli
model~$(\mu, \sigma)$ derived via analytical marginalization over continuous
Cartesian shifts, PROBE retains grid-based efficiency while
improving cross-session robustness.
Density-adaptive scaling prevents over-smoothing on sparse sensors,
and multiplicative fusion of the Bernoulli-KL Jaccard and FFT height
cosine similarity yields a discriminative and {robust} pipeline.
{
A sweep across four sensor types (16--128 channels) confirms
that $\sigma_t{=}2.0$\,m lies within a broad performance plateau
(Fig.~\ref{fig:sigma_sensitivity}), establishing a sensor-independent default.}

{
The shrinkage mechanism suppresses symmetric boundary
noise, but provides limited protection against \emph{asymmetric}
occupancy changes (e.g., parked vehicles present only in the database);
dynamic-object filtering or temporal priors are potential mitigations.}
{Similarly,} performance degrades on low-channel sensors (e.g., 16-beam)
{where wide vertical gaps induce artificial structural
voids; dynamically increasing cell size proportionally to the
sparsity may restore the minimum occupancy density for stable
probabilistic modeling.}
{More broadly,} like all BEV descriptors, PROBE remains vulnerable to extreme
translational offsets ($> \SI{5}{\metre}$).
{
While the Bernoulli formulation intrinsically supports
restricted-FOV LiDARs via $\sigma$-gating, missing sectors
distort the retrieval keys, motivating future work on
uncertainty-aware keys, asymmetric mapping, and distributional
height encodings to unify geometric and learned representations.}

\bibliographystyle{IEEEtran}

\end{document}